\documentclass[journal]{IEEEtran/IEEEtran}

\ifCLASSOPTIONcompsoc
  \usepackage[caption=false,font=normalsize,labelfont=sf,textfont=sf]{subfig}
\else
 \usepackage[caption=false,font=footnotesize]{subfig}
\fi

\usepackage{amssymb} 
\usepackage{booktabs, multicol, multirow, bigstrut, array}
\newcolumntype{R}[1]{>{\raggedleft\let\newline\\\arraybackslash\hspace{0pt}}m{#1}}
\usepackage{hhline}
\usepackage{cite}

\usepackage[dvips]{graphicx}
 \DeclareGraphicsExtensions{.eps,.bmp}
 \usepackage{pgf}
 \usepackage{tikz}
 \usepackage{pstricks} 

\usepackage[cmex10]{amsmath}
\usepackage{fixmath}
\usepackage{fixltx2e}

\usepackage[flushleft]{threeparttable}

\begin{document}

\title{Deep CNNs along the Time Axis with Intermap Pooling for Robustness to Spectral Variations}

\author{Hwaran~Lee,~
        Geonmin~Kim,~
        Ho-Gyeong~Kim,~
        Sang-Hoon~Oh,~
        and~Soo-Young~Lee%
         
\thanks{This work was supported by the ICT R\& D program of MSIP \/ IITP. [R0126-15-1117, Core technology development of the spontaneous speech dialogue processing for the language learning]}%
\thanks{
H. Lee, G. Kim, H.-G. Kim and S.-Y. Lee are with the School of Electrical Engineering, Korea Advanced Institute of Science and Technology, Daejeon 305-701, Korea. (e-mail:
\{hwaran.lee, gmkim90, hogyeong, sylee\}@kaist.ac.kr)}
\thanks{
S.-H Oh is with the Division of Information and Communication Convergence Engineering, Mokwon University, Daejeon, 302-318, Korea. (e-mail: shoh@mokwon.ac.kr)}%
}

\maketitle

\begin{abstract}
Convolutional neural networks (CNNs) with convolutional and pooling operations along the frequency axis have been proposed to attain invariance to frequency shifts of features. However, this is inappropriate with regard to the fact that acoustic features vary in frequency. In this paper, we contend that convolution along the time axis is more effective. We also propose the addition of an intermap pooling (IMP) layer to deep CNNs. In this layer, filters in each group extract common but spectrally variant features, then the layer pools the feature maps of each group. As a result, the proposed IMP CNN can achieve insensitivity to spectral variations characteristic of different speakers and utterances. The effectiveness of the IMP CNN architecture is demonstrated on several LVCSR tasks. Even without speaker adaptation techniques, the architecture achieved a WER of 12.7\% on the SWB part of the Hub5\rq2000 evaluation test set, which is competitive with other state-of-the-art methods.
\end{abstract}


\IEEEpeerreviewmaketitle

\section{Introduction}
\label{sec:intro}
\IEEEPARstart{A}{coustic} modeling with deep learning has demonstrated remarkable performance improvements in automatic speech recognition \cite{Mohamed2010, Dahl2011,  Lee2011, Hinton2012}. Deep neural networks (DNNs) are trained to label each frame of processed speech data with the state of a hidden Markov model (HMM). However, there is a difficulty due to the fact that acoustic features vary widely in frequency and articulation rate depending on harmonics of the vocal tract and characteristic speaking styles. 

Efforts to effectively handle these variations can be categorized into feature-level and model-level approaches. Amongst feature-level approaches, speaker-adapted methods such as fMLLR \cite{Rath2013} have been proposed. Acoustic features concatenated with i-vectors, which represent speaker information, also have been employed as input for DNNs \cite{Saon2013, Senior2014}. Model-level approaches have employed hybrid NN-HMM systems with convolutional neural networks (CNNs) \cite{Abdel-Hamid2012, Abdel-hamid2013a, Abdel-Hamid2014} and recurrent neural networks (RNNs) \cite{Graves2013a, Graves2013, Beaufays2014}.

In particular, CNNs have advantages in terms of capturing local features through weight sharing while remaining robust to slight translations of these features through pooling. Structural advantages of CNNs enable the modeling of speech data without feature-level engineering, such as spectrograms or mel filter-banks. Previous researchers introduced time-delayed neural networks (TDNNs), which are CNNs with convolutions along the time axis to learn the temporal dynamics of features  \cite{ Waibel1989, Lee2009a, Peddinti2015}. Other researchers have applied convolutions along the frequency axis to attain invariance to frequency-shifts \cite{Abdel-Hamid2012, Sainath2013b}. However, acoustic features of speech vary in frequency, so that weight sharing along the frequency axis may not be appropriate. The limited weight sharing method in which weights are convolved only within a subsection of frequency-bands has been employed in efforts to overcome this problem \cite{Abdel-Hamid2014}. However, settling on appropriate band divisions and filters will require further work.

One limitation common to the preceding approaches is that most of them have employed only one or two convolution and pooling layers. Another limitation is that the relationship or topography of filters trained in supervised learning has not been intensively investigated. For unsupervised feature extraction, previous researchers imposed sparsity terms over small groups or neighborhoods in feature maps of image \cite{Hyvarinen2000, Hyvarinen2001, Kavukcuoglu2009} and speech data \cite{Kim2005, Terashima2012}. They attained topographically-organized maps of smoothly varying oriented edge filters or tonotopic disordered topography of spectrotemporal features, such as those found in the primary visual or auditory cortex (V1, A1) respectively.

In this paper, we argue that convolution along the time axis is more effective than along the frequency axis for acoustic models. In order that the network learns temporal dynamics adequately, we increase the depth of convolution layers that have small filters. Instead of frequency-axis convolution and pooling, we propose the addition of a convolutional maxout layer, namely an {\it intermap pooling} (IMP) layer in order to increase robustness to spectral variations. Previously, a convolutional maxout network has been proposed \cite{Cai2014}, however, it applied convolutions along the frequency axis. We show that the IMP CNNs with the time convolution reduce the word error rates more. As a result, the IMP CNNs can both model temporal dynamics and remain robust to spectral variations.

\section{Convolution Neural Networks}
\label{ssec:CNNs}
CNNs consist of the alternation of convolution and pooling layers, and fully connected layers in the top-most layer. 
Let $ \mathbf{H}^{(l)}$ stand for input to the $l$th convolution layer having $K$ filters, with the $k$th convolution filter denoted $ \mathbf{W}^{(l)}_k \in \mathbb{R}^{M \times N \times G }$ with $M$ and $N$ denoting a filter’s height and width respectively, and $G$ designating the number of feature maps of the input. 
A bias term $b^{(l)}_k$ is shared inside the $k$th feature map. 
Thus, from any input $ \mathbf{H}^{(l)}$, the output $ \mathbf{\tilde{H}}^{(l+1)}  \in \mathbb{R}^{I \times J \times K} $ can be calculated as
\begin{equation}
\footnotesize
\begin{split}
  \mathbf{\tilde{H}}^{(l+1)}_{(i,j,k)} = f( ( \mathbf{H}^{(l)} \ast \mathbf{W}^{(l)}_k   )_{(i,j)}+b^{(l)}_k  )  \\
\footnotesize{ (  \text{for  } 1 \leq i \leq I, 1 \leq j \leq J)},
\end{split}
 \end{equation}
where $I$ and $J$ are height and width of each output feature map, and $f$ is a non-linear function such as sigmoid or rectified linear unit (ReLU).

An intramap pooling layer, typically called ``max-pooling'', propagates the maximum value from each sub-region in each feature map. For non-overlapping sub-regions with height $p$ and width $q$, the output from this pooling layer is given by
\begin{equation}
\footnotesize
 \mathbf{H}^{(l)}_{(i,j,k)} = \max_{ \substack{ \alpha = -p+1, ... , 0 \\
    \beta=-q+1 ,...,0 } }
     \mathbf{\tilde{H}}^{(l)}_{(ip+\alpha, jq+\beta, k)} .
 \end{equation}
Intramap pooling layers have blurring effects on feature maps, with the result being that the CNN is more robust to locally translated features.

\section{Intermap Pooling Layers}
\label{sec:Intermap pooling layers}

There are several categories of acoustic features such as harmonics, formants, and on/offsets (i.e., start and end points of speech). Spectral variations of acoustic features appear as shifts in the frequency axis over time (spectro-temporal modulation). In order to ensure the robustness of our model to spectral variations, we propose the addition of a convolutional maxout layer, the {\it intermap pooling} (IMP) layer. Like the maxout networks \cite{Goodfellow2013}, this layer {\it groups} the filters, and {\it pools} the feature maps inside a group.  

\begin{figure}[t]
 \centering
  \includegraphics[width=7.5cm]{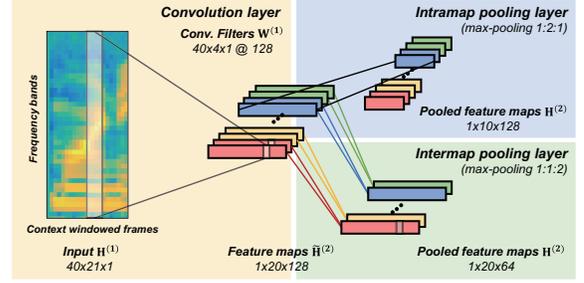}
\caption{ An illustration of a convolution layer followed by an intramap pooling layer or an intermap pooling layer. 
The sizes of convolution filters and feature maps are denoted as \lq(the freq. axis)$\times$(the time axis)$\times$(the feature axis)'.
The number of filters is denoted after \lq @'.
The pooling size is denoted as \lq(the freq. axis):(the time axis):(the feature axis)'.
The convolution input is padded with zeros at both ends in the time axis in order to preserve frame length after convolutions.}
 \label{fig:architecture}
\end{figure}

Specifically, an intermap pooling layer partitions feature maps into a set of groups. Then each group propagates the maximum activation value at each position. Formally, the output of the $k$th group consisting of $r$ consecutive feature maps is given by
\begin{equation}
\footnotesize
  {\mathbf{H}}^{(l)}_{(i,j,k)} = \max_{  \gamma = -r+1, ... , 0 }
     \mathbf{\tilde{H}}^{(l)}_{(i, j, kr + \gamma )} .
 \end{equation}
The structural comparison of intermap and intramap pooling layers is shown in Fig.\ref{fig:architecture}. Note that the method pursued in this paper does not introduce any additional learning terms except for the intermap grouping of filters.

The central idea to the IMP CNN is that the filters in each group learn common but spectrally variant features, such as frequency-shifted harmonics, and the pooled feature map is invariant to those feature variations within the group.
The pooled feature maps $ \mathbf{H}^{(l)} $  are representative of the feature maps in each group. Through supervised learning, the pooled feature maps become discriminative of features for recognizing phonemes. 
{\it
For a phoneme, since spectral variations among different speakers and utterances are not discriminative information, the individual filters in a group spontaneously represent common but spectrally variant features, even though the layer does not ensure this.}

\section{Deep CNN Architecture \\ for Acoustic Modeling}
\label{sec: Acoustic modeling with a very deep CNN }

Since short-term temporal dynamics are shared within every frame of a given speech sample, sharing filters along the time axis is reasonable. However, sharing filters along the frequency axis may not be suitable, because features within lower frequency-band regions are significantly different from those in the higher regions. Instead of convolution along the frequency axis, our architecture employs an intermap pooling layer following the first convolution layer. This approach demonstrates robustness not only to frequency-shifted features but also to spectro-temporally distorted features. Moreover, it does not require engineered efforts to consider the varying characteristics of different frequency-bands.

A sufficiently {\it deep} depth of convolution and pooling layers is necessary to precisely represent complex acoustic features with temporal and spectral variations. Individual frames also should be labeled as minutely as the number of HMM states, which is more than thousands in tri-phone modeling. However, context windowed inputs are too tiny (e.g., 21 frames) and stacking multiple intramap pooling layers decreases the feature map size in proportion to the pooling size, thereby restricting the depth of CNNs. Since, previous researchers have chosen large convolution filter and intramap pooling sizes, sufficient increases in depths of CNNs have not been realized.

As illustrated in Fig.\ref{fig:architecture}, the IMP CNN architecture applies convolution and intramap pooling layers only along the time axis. The pooling size of the intramap pooling layers is small so that it does not decrease temporal resolution much. Furthermore, motivated by the performance of very deep CNNs \cite{Simonyan2015}, we inserted convolution layers with small filters (of size 1x3) between two intramap pooling layers. The combination of filters before pooling layers increases non-linearity, and this results in a network that has rich feature expressions.

\begin{figure}[!t]
 \centering
  \includegraphics[width=6.5cm]{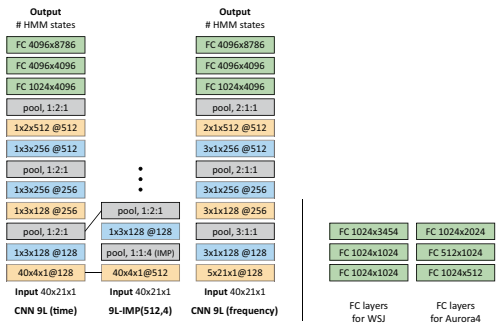}
\caption{ Configurations of CNN architectures for SWBD. The  ReLU non-linearity function is used on the top of every activation. We made 6, 9, 12, and 15 layers by excluding or repeating the blue colored layers. For WSJ and Aurora4 datasets, the fully connected layers have smaller hidden neurons, and remainder are the same. }
 \label{fig:configure}
\end{figure}


 \section{Experimental Results}
\label{sec: Experimental Results }
\subsection{Experiments setup}
\label{ssec: experiment setup}

We conducted experiments using the 300 hour Switchboard-I Release 2 (SWBD) dataset \cite{Godfrey1992} which is conversational telephone speech task as well as the Wall Street Journal (WSJ) corpus \cite{Paul1994} and Aurora4 database which are read speech. We used the 81-hour training dataset (SI-284) of the WSJ corpus. The Aurora4 database is a subset of the WSJ in which clean utterances are added with different noise types and/or convolved with microphone distortions. The following results are for the trained IMP CNN on the multi-conditioned training dataset.

The raw speech signal is processed via short-time Fourier transform (STFT) with a 25ms Hamming window and 10ms window shifts. We used 40-dimensional log-mel filter bank features without the energy coefficient, and concatenated frames with a context window size of 21 ($\pm$10 frames) to feed them into networks as inputs. We trained the GMM-HMM system over fMLLR features. The forced alignment of each frame by the GMM-HMM baseline system is the target label of the neural networks.

After random initialization of weights and biases from the Gaussian distribution $\mathcal{N}$(0,0.01) and $\mathcal{N}$(0,0.5) respectively, the CNNs were optimized by the stochastic gradient descent (SGD) method.
In particular, for CNNs deeper than 9-layers, we faced with infeasible training, because each layer back-propagates errors by multiplying its small initial weights, resulting in vanishing gradients. Therefore, we increased the standard deviation ($\sigma$) of the Gaussian distribution in lower layers.
Each layer is trained with a momentum of 0.9, an L2-decay term of 0.0005, and mini-batch size of 512. After one epoch of training, the trained model is accepted if the validation cost decreases. Otherwise, the trained model is rejected and training starts again from the latest accepted model with a halved learning rate. The initial learning rate is 0.01, and the training stops after 50 epochs. Our implementation is developed upon the KALDI toolkit \cite{Povey2011}.

For the SWBD task, we decode speech using a trigram language model (LM) of 30k vocabularies which is trained on 3M words, and then we rescore the decoding results using 4-gram LM which is trained on Fisher English Part 1 transcripts \cite{Cieri2004}. For the WSJ and Aurora4 corpus, we used a 146K word extended dictionary and the trigram pruned language model which is exactly the same as the \lq s5 \rq recipe in the KALDI.

\begin{figure}[!t]
 \centering
  \includegraphics[width=8cm]{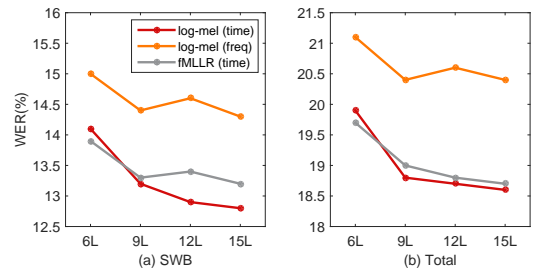}
\caption{ Decoding results of CNNs on the Switchboard evaluation sets (Hub5'2000). CNNs with different convolution axis (time, frequency) and input features (log-mel, fMLLR) are compared.  (a) Switchboard subset (b) Total (SWB and CallHome).}
 \label{fig:depth_swbd}
\end{figure}

\subsection{Convolution axis and depth of CNNs}
\label{ssec:Very deep CNNs}
Fig. \ref{fig:depth_swbd} shows the decoding results of CNNs on SWBD evaluation sets with various depths, from 6 layers up to 15 layers (configurations are described in Fig. \ref{fig:configure}). Deeper CNNs produced lower WERs, with the 15-layer CNN achieving a WER of 12.8\% for SWB and 18.6\% for total evaluation sets. Moreover, it is validated that convolution along the time axis always outperforms convolution along the frequency axis. Furthermore, CNNs trained over log-mel features had lower WER as fMLLR features when CNN has more than 9 layers. These results show that weight sharing along the time axis more effectively reduces the WER, and that increased non-linearity obviates preprocessing for speaker adaptation.

\begin{table}[!t]
  \centering
  \caption{WERs(\%) of CNNs with an Intermap Pooling Layer (IMP) and an Overlapping IMP (IMPO) on SWBD.}
  \small
    \begin{tabular}{l|c|c}
    \hline
    \multicolumn{1}{c|}{Network} & SWB & Total \bigstrut\\
    \hline
    \hline
    9L & 13.2 & 18.8 \bigstrut\\
    \hline
    9L-IMP(128,2) & 13.2 & 18.8 \bigstrut\\
    9L-IMP(256,2) & 13.2 & 18.7 \bigstrut\\
    9L-IMP(512,4) & \textbf{12.7} & \textbf{18.5} \bigstrut\\
    9L-IMP(768,6) & 12.9 & 18.7 \bigstrut\\
    \hline
     9L-IMPO(512,4) & 13.0 & 18.6 \bigstrut\\
    \hline
    \end{tabular}%
  \label{tab:table_cnn_imp_swbd}%
\end{table}%

\renewcommand{\multirowsetup}{\centering}
\begin{table}[!t]
  \centering
  \caption{Intermap Pooling in different convolution axis.}
    \begin{tabular}{r|c|c|c|c}
    \hline
      & \multicolumn{2}{c|}{SWB} & \multicolumn{2}{c}{Total} \bigstrut[t]\\
      & 9L & 9L-IMP & 9L & 9L-IMP \bigstrut[b]\\
    \hline
    \hline
    \multicolumn{1}{c|}{log-mel (time)} & 13.2 & \textit{\underline{12.7}} & 18.8 & \textit{\underline{18.5}} \bigstrut[t]\\
    \multicolumn{1}{c|}{log-mel (freq.)} & \textit{\underline{14.4}} & 14.7 & \textit{\underline{20.4}} & 20.7 \\
    \multicolumn{1}{c|}{fMLLR (time)}  & \textit{\underline{13.3}} & 13.5 & \textit{\underline{19.0}} & 19.1 \bigstrut[b]\\
    \hline
    \end{tabular}%
  \label{tab:cnn_imp_compare}%
\end{table}%

\subsection{IMP CNNs}
\label{ssec:Intermap pooling layers}
Decoding results of IMP-CNNs with different numbers of maps and pooling sizes are compared in Table \ref{tab:table_cnn_imp_swbd}. 
We further investigated an intermap pooling layer in which groups overlap (IMPO) each other with a stride of one.
All CNNs with an intermap pooling layer performed better than the 9-layer CNN without. Especially, the `9L-IMP(512, 4)' CNN performs the best with a WER of 12.7\% for SWB test set, showing a 3.78\% relative improvement over the 9-layer CNN. 
Note Table \ref{tab:cnn_imp_compare} that when the IMP layer is applied to CNNs along the frequency axis or over fMLLR features, performance declines. In addition, IMP CNN performed well on the WSJ and Aurora4 corpus as shown in Table\ref{tab:table_cnn_imp_wsj} and \ref{tab:table_cnn_imp_aurora4},  respectively. It is remarkable that IMP layers contribute robustness to spectral variations in both clean and noisy conditions.

\begin{table}[!t]
  \centering
  \caption{WERs(\%) of CNNs with an Intermap Pooling Layer  on WSJ.}
    \begin{tabular}{l|c|c|c}
    \hline
      & 9L & 9L-IMP & Rel. (\%) \bigstrut\\
    \hline
    \hline
    Eval'92 & 4.27 &\textbf{3.93} & 7.96 \bigstrut\\
    \hline
    \end{tabular}%
  \label{tab:table_cnn_imp_wsj}%
\end{table}%

\begin{table}[!t]
  \centering
  \caption{WERs(\%) of CNNs with an Intermap Pooling Layer  on Aurora4.}
  
    \begin{tabular}{l|c|c|c|c|c}
    \hline
      & \multicolumn{1}{c|}{Clean} & \multicolumn{1}{c|}{Noise} & \multicolumn{1}{c|}{Channel} & \multicolumn{1}{c|}{Channel+Noise} & \multicolumn{1}{c}{Avg.} \bigstrut\\
    \hline
    \hline
    9L & 3.36 & 7.05 & 8.14 & 18.05 & 11.58 \bigstrut\\
   9L-IMP & \textbf{3.14} & \textbf{6.64} & \textbf{7.86} & \textbf{17.88} & \textbf{11.29} \bigstrut\\
    \hline
    Rel. (\%) & 6.55 & 5.82 & 3.44 & 0.92 & 2.50 \bigstrut\\
    \hline
    \end{tabular}%
  \label{tab:table_cnn_imp_aurora4}%
\end{table}%

\subsection{Analysis on learnt filters}
\label{ssec: Analysis on acoustic features }
\begin{figure}[]
 \centering
  \includegraphics[width=6.9cm]{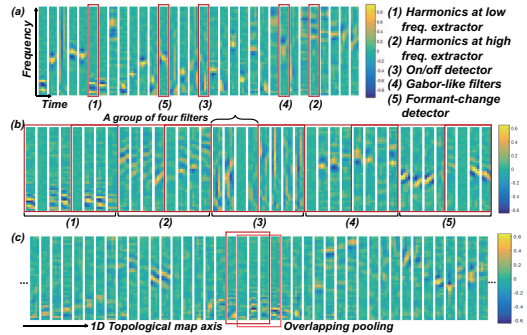} 
 \caption{ The trained filters of the first convolution layer.
(a) Filters of the 9-layer CNN. The top 22 filters that have the biggest L2-norm are sorted in decreasing order.
(b) 10 groups of four filters in \lq 9L-IMP(512,4)'. The order of groups (red boxes) is arranged according to the feature category of each group.
(c) A part of successive filters in `9L-IMPO(512,4)'.}
\label{fig:features}
\end{figure}

Learnt filters of the first convolution layer are visualized in Fig.\ref{fig:features} (a). There are {\it five categories of spectrotemporal features} in the filters. (1) Harmonic features are narrow in the low frequency-region and (2) broad in the high frequency-region. (3) The on/off-set detecting filters are temporally selective, but are also sensitive to several frequencies. (4) The features of Gabor-like filters are centered on some frequency-bands, which presumably detect formants. (5) The features of formant changes are directional diagonal lines, spectrotemporal modulations, in the middle frequency-bands. Note that different features appear in different frequency-bands, and that local features of each type have different bandwidth sizes. 

The trained filters in each group of the intermap pooling layer are presented in Fig.\ref{fig:features} (b). Importantly, most filters in a group belong to a common category. 
For example, the filters in harmonic extractor and formant change detector groups have marginally shifted features on the frequency axis. 
This figure verifies that the intermap pooling lead the filters of a group to extract common but spectrally variant features, although there are no additional architectural constraints to guarantee this. 

The consecutive trained filters of the IMPO layer are drawn in Fig. \ref{fig:features}(c). The filters form a 1-dimensional topological map, where neighboring filters respond to similar spectrotemporal features.
Along the topological map axis, filters appear discontinuously between feature categories, reflecting the fact that feature categories become definitely distinguishable to the system as it is trained. 
In recent neurophysiological studies, there is consensus that multiple tonotopic maps exist in the human auditory system \cite{Saenz2014}. However, few studies suggest that this topography includes other sound features, such as temporal, spectral, and joint modulations \cite{Barton2012, Herdener2013, Santoro2014}. 
The trained topography may provide a clue as to how human auditory neurons organize to efficiently process information in A1.

%
%

  


\begin{table}[!t]
  \centering
  \caption{Comparison of WERs(\%) for Different Models on SWBD.}
    \begin{tabular}{c|l|c|c}
 \hline
    Input features & \multicolumn{1}{c|}{Model} & SWB & Total \bigstrut\\
    \hline
    \hline
        \multirow{2}[2]{*}{fMLLR} & \multicolumn{1}{l|}{GMM-HMM} & 19.5 & 26.7 \bigstrut[t]\\
      & \multicolumn{1}{l|}{Maxout 7L} & 14.2 & 20.0 \bigstrut[b]\\
    \hline
    \multirow{4}[2]{*}{log-mel filterbanks} & \multicolumn{1}{l|}{Maxout 7L} & 14.6 & 20.7 \bigstrut[t]\\
      & \multicolumn{1}{l|}{CNN 9L} & 13.2 & 18.8 \\
      & \multicolumn{1}{l|}{CNN 9L-IMP(512,4)} & \textit{\underline{12.7}} & \textbf{18.5} \\
      & \multicolumn{1}{l|}{CNN 15L} & 12.8 & 18.6 \bigstrut[b]\\
    \hline
    MFCC + i-vectors & \multicolumn{1}{l|}{TDNN 4L \cite{Peddinti2015}} & 12.9 & 19.2 \bigstrut\\
    \hline
    \multirow{2}[2]{*}{VTL-warped log-mel} & CNN 8L (2conv+6fc)  \cite{Saon2015}& \textbf{12.6} & - \bigstrut[t]\\
      & CNN 13L (10conv+3fc)  \cite{Sercu2016a}& \textbf{11.8} & - \bigstrut[b]\\
    \hline
    \end{tabular}%
  \label{tab:compare_swbd}%
\end{table}%


\subsection{Comparison of the IMP CNN}
\label{ssec: Comparison of the proposed CNNs with Other Models}

For comparison, we trained max-out networks that have 7 layers with 2,000 hidden neurons and 400 groups on both fMLLR and filter-bank features. The comparison of the decoding results is summarized in Table \ref{tab:compare_swbd}. The \lq 9L-IMP(512, 4)\rq IMP CNN improved on the GMM-HMM baseline (19.5\%) and the max-out network (14.6\%), demonstrating a 34.87\% and 13.01\% relative improvement respectively. Also, it performs on par with a 15-layer CNN, i.e. a non-IMP CNN with six additional convolution layers. 
Finally, the IMP CNN is compared with the TDNN \cite{Peddinti2015} and the CNNs which employed 2-dimensional convolutions \cite{Saon2015, Sercu2016a}. Note that we only compare other previous results without any sequence training such as sMBR. Even though our deep CNN did not use any speaker adaptation techniques, it yielded a comparative word error rate simply by employing intermap pooling and by increasing depths.

\section{Conclusion}
\label{sec:conclusion}

In this paper, the present experiments demonstrate that convolution along the time axis is more effective than along the frequency axis when processing speech. Depth in convolution layers is crucial for the sufficient representation of the complex temporal dynamics inherent in the acoustic features of speech. In order to achieve greater robustness to spectral variations in speech recognition, we proposed the addition of intermap pooling (IMP) to CNNs. Through visualization of the trained filters, we verified that filters grouped together learn similar spectrotemporal features and form a topological map. In the end, even without any speaker adaptation techniques, the proposed IMP CNN delivered competitive performance on the Switchboard, WSJ, and Aurora4 databases.

\ifCLASSOPTIONcaptionsoff
  \newpage
\fi

\bibliographystyle{IEEEtran}

\newpage
\bibliography{references.bib}

\begin{thebibliography}{10}
\providecommand{\url}[1]{#1}
\csname url@samestyle\endcsname
\providecommand{\newblock}{\relax}
\providecommand{\bibinfo}[2]{#2}
\providecommand{\BIBentrySTDinterwordspacing}{\spaceskip=0pt\relax}
\providecommand{\BIBentryALTinterwordstretchfactor}{4}
\providecommand{\BIBentryALTinterwordspacing}{\spaceskip=\fontdimen2\font plus
\BIBentryALTinterwordstretchfactor\fontdimen3\font minus
  \fontdimen4\font\relax}
\providecommand{\BIBforeignlanguage}[2]{{%
\expandafter\ifx\csname l@#1\endcsname\relax
\typeout{** WARNING: IEEEtran.bst: No hyphenation pattern has been}%
\typeout{** loaded for the language `#1'. Using the pattern for}%
\typeout{** the default language instead.}%
\else
\language=\csname l@#1\endcsname
\fi
#2}}
\providecommand{\BIBdecl}{\relax}
\BIBdecl

\bibitem{Mohamed2010}
A.-R. Mohamed, G.~E. Dahl, and G.~Hinton, ``{Acoustic Modeling using Deep
  Belief Networks},'' \emph{Audio, Speech, Lang. Process. IEEE Trans.},
  vol.~20, no.~1, pp. 14--22, 2010.

\bibitem{Dahl2011}
G.~Dahl, D.~Yu, L.~Deng, and A.~Acero, ``{Large vocabulary continuous speech
  recognition with context-dependent DBN-HMMs},'' \emph{INTERSPEECH}, pp.
  4688--4691, 2011.

\bibitem{Lee2011}
J.~Lee and S.-Y. Lee, ``{Deep learning of speech features for improved phonetic
  recognition},'' \emph{Acoust. Speech Signal Process. (ICASSP), 2011 IEEE Int.
  Conf. on.}, no. August, pp. 1249--1252, 2011.

\bibitem{Hinton2012}
G.~Hinton, L.~Deng, D.~Yu, G.~Dahl, A.-R. Mohamed, N.~Jaitly, V.~Vanhoucke,
  P.~Nguyen, T.~Sainath, and B.~Kingsbury, ``{Deep Neural Networks for Acoustic
  Modeling in Speech Recognition},'' \emph{Signal Process. Mag.}, vol.~29,
  no.~6, pp. 82--97, 2012.

\bibitem{Rath2013}
S.~P. Rath, D.~Povey, K.~Vesel, and J.~H. Cernock, ``{Improved feature
  processing for Deep Neural Networks},'' \emph{INTERSPEECH}, pp. 1--5, 2013.

\bibitem{Saon2013}
G.~Saon, H.~Soltau, D.~Nahamoo, and M.~Picheny, ``{Speaker adaptation of neural
  network acoustic models using i-vectors},'' \emph{2013 IEEE Work. Autom.
  Speech Recognit. Underst.}, pp. 55--59, 2013.

\bibitem{Senior2014}
A.~Senior and I.~Lopez-Moreno, ``{Improving DNN Speaker Independence With
  I-Vector Inputs},'' \emph{2014 IEEE Int. Conf. Acoust. Speech Signal
  Process.}, pp. 225--229, 2014.

\bibitem{Abdel-Hamid2012}
O.~Abdel-Hamid, A.-R. Mohamed, H.~Jiang, and G.~Penn, ``{Applying convolutional
  neural networks concepts to hybrid NN-HMM model for speech recognition},''
  \emph{IEEE Int. Conf. Acoust. Speech Signal Process.}, 2012.

\bibitem{Abdel-hamid2013a}
O.~Abdel-Hamid, L.~Deng, and D.~Yu, ``{Exploring Convolutional Neural Network
  Structures and Optimization Techniques for Speech Recognition},''
  \emph{INTERSPEECH}, no. August, pp. 3366--3370, 2013.

\bibitem{Abdel-Hamid2014}
O.~Abdel-Hamid, A.-R. Mohamed, H.~Jiang, L.~Deng, G.~Penn, and D.~Yu,
  ``{Convolutional Neural Networks for Speech Recognition},'' \emph{Audio,
  Speech, Lang. Process. IEEE/ACM Trans.}, vol.~22, no.~10, pp. 1533--1545,
  2014.

\bibitem{Graves2013a}
A.~Graves, A.-R. Mohamed, and G.~Hinton, ``{Speech Recognition With Deep
  Recurrent Neural Networks},'' \emph{Acoust. Speech Signal Process. (ICASSP),
  2013 IEEE Int. Conf. on. IEEE}, no.~3, pp. 6645--6649, 2013.

\bibitem{Graves2013}
A.~Graves, N.~Jaitly, and A.~R. Mohamed, ``{Hybrid speech recognition with Deep
  Bidirectional LSTM},'' \emph{Autom. Speech Recognit. Underst. (ASRU), 2013
  IEEE Work. on. IEEE}, pp. 273--278, 2013.

\bibitem{Beaufays2014}
F.~Beaufays, H.~Sak, and A.~Senior, ``{Long Short-Term Memory Recurrent Neural
  Network Architectures for Large Scale Acoustic Modeling Has},''
  \emph{INTERSPEECH}, no. September, pp. 338--342, 2014.

\bibitem{Waibel1989}
A.~Waibel, T.~Hanazawa, G.~E. Hinton, K.~Shikano, and K.~J. Lang, ``{Phoneme
  recognition using time-delay neural networks},'' \emph{Acoust. Speech Signal
  Process. IEEE Trans.}, vol.~37, no.~3, pp. 328--339, 1989.

\bibitem{Lee2009a}
H.~Lee, P.~Pham, Y.~Largman, and A.~Ng, ``{Unsupervised feature learning for
  audio classification using convolutional deep belief networks.}'' \emph{Adv.
  Neural Inf. Process. Syst.}, pp. 1--9, 2009.

\bibitem{Peddinti2015}
V.~Peddinti, D.~Povey, and S.~Khudanpur, ``{A time delay neural network
  architecture for efficient modeling of long temporal contexts},''
  \emph{INTERSPEECH}, pp. 2--6, 2015.

\bibitem{Sainath2013b}
T.~N. Sainath, A.-R. Mohamed, B.~Kingsbury, and B.~Ramabhadran, ``{Deep
  Convolutional neural networks for LVCSPR},'' \emph{Acoust. Speech Signal
  Process. (ICASSP), 2013 IEEE Int. Conf. on.}, pp. 10--14, 2013.

\bibitem{Hyvarinen2000}
A.~Hyv{\"{a}}rinen and H.~Patrik, ``{Emergence of phase and shift invariant
  features by decomposition of natural images into independent feature
  subspaces},'' \emph{Neural Comput.}, vol.~12, no.~7, pp. 1705----1720, 2000.

\bibitem{Hyvarinen2001}
A.~Hyv{\"{a}}rinen, P.~O. Hoyer, and M.~Inki, ``{Topographic independent
  component analysis},'' \emph{Neural Comput.}, vol.~13, no.~7, pp. 1527--1558,
  2001.

\bibitem{Kavukcuoglu2009}
K.~Kavukcuoglu, M.~Ranzato, R.~Fergus, and Y.~Le-Cun, ``{Learning Invariant
  Features through Topographic Filter Maps},'' \emph{Comput. Vis. Pattern
  Recognit.}, pp. 1605--1612, 2009.

\bibitem{Kim2005}
T.~Kim and S.-Y. Lee., ``{Learning self-organized topology-preserving complex
  speech features at primary auditory cortex},'' \emph{Neurocomputing}, vol.
  65-66, pp. 793--800, 2005.

\bibitem{Terashima2012}
H.~Terashima and M.~Okada, ``{The topographic unsupervised learning of natural
  sounds in the auditory cortex},'' \emph{Adv. Neural Inf. Process. Syst.}, pp.
  1--9, 2012.

\bibitem{Cai2014}
M.~Cai, Y.~Shi, J.~Kang, J.~Liu, and T.~Su, ``{Convolutional maxout neural
  networks for low-resource speech recognition},'' \emph{Proc. 9th Int. Symp.
  Chinese Spok. Lang. Process. ISC SLP 2014}, pp. 133--137, 2014.

\bibitem{Goodfellow2013}
I.~J. Goodfellow, D.~Warde-Farley, M.~Mirza, A.~Courville, and Y.~Bengio,
  ``{Maxout Networks},'' \emph{arXiv Prepr.}, pp. 1319--1327, 2013.

\bibitem{Simonyan2015}
K.~Simonyan and A.~Zisserman, ``{Very Deep Convolutional Networks for
  Large-Scale Image Recoginition},'' \emph{arXiv Prepr.}, pp. 1--14, 2015.

\bibitem{Godfrey1992}
J.~J. Godfrey, E.~C. Holliman, and J.~McDaniel, ``{SWITCHBOARD telephone speech
  corpus for research and development},'' \emph{Proc. 1992 IEEE Int. Conf.
  Acoust. Speech, Signal Process.}, vol.~1, pp. 517--520, 1992.

\bibitem{Paul1994}
D.~B. Paul and J.~M. Baker, ``{The Design for the Wall Street Journal-based CSR
  Corpus},'' \emph{Proc. Work. Speech Nat. Languae}, 1994.

\bibitem{Povey2011}
D.~Povey, A.~Ghoshal, G.~Boulianne, L.~Burget, O.~Glembek, N.~Goel,
  M.~Hannemann, P.~Motlicek, Y.~Qian, P.~Schwarz, J.~Silovsky, G.~Stemmer, and
  K.~Vesely, ``{The kaldi speech recognition toolkit},'' \emph{Autom. Speech
  Recognit. Underst. (ASRU), 2011 IEEE Work. on. IEEE}, pp. 1--4, 2011.

\bibitem{Cieri2004}
C.~Cieri, D.~Miller, and K.~Walker, ``{The Fisher corpus: a Resource for the
  Next Generations of Speech-to-Text},'' \emph{Proc. Lr.}, vol.~4, pp. 69--71,
  2004.

\bibitem{Saenz2014}
M.~Saenz and D.~R. Langers, ``{Tonotopic mapping of human auditory cortex},''
  \emph{Hear. Res.}, vol. 307, pp. 42--52, 2014.

\bibitem{Barton2012}
B.~Barton, J.~H. Venezia, K.~Saberi, G.~Hickok, and A.~A. Brewer, ``{Orthogonal
  acoustic dimensions de fi ne auditory fi eld maps in human cortex},''
  \emph{Proc. Natl. Acad. Sci.}, 2012.

\bibitem{Herdener2013}
M.~Herdener, F.~Esposito, K.~Scheffler, P.~Schneider, N.~K. Logothetis,
  K.~Uludag, and C.~Kayser, ``{Spatial representations of temporal and spectral
  sound cues in human auditory cortex},'' \emph{Cortex}, vol.~49, no.~10, pp.
  2822--2833, 2013.

\bibitem{Santoro2014}
R.~Santoro, M.~Moerel, F.~{De Martino}, R.~Goebel, K.~Ugurbil, E.~Yacoub, and
  E.~Formisano, ``{Encoding of Natural Sounds at Multiple Spectral and Temporal
  Resolutions in the Human Auditory Cortex},'' \emph{PLoS Comput. Biol.},
  vol.~10, no.~1, p. e1003412, 2014.

\bibitem{Saon2015}
G.~Saon, H.-K.~J. Kuo, S.~Rennie, and M.~Picheny, ``{The IBM 2015 English
  Conversational Telephone Speech Recognition System},'' \emph{Interspeech
  2015}, 2015.

\bibitem{Sercu2016a}
T.~Sercu, C.~Puhrsch, B.~Kingsbury, and Y.~LeCun, ``{Very deep multilingual
  convolutional neural networks for LVCSR},'' pp. 2--6, 2016.

\end{thebibliography}
\label{sec:ref}

\end{document}